\documentclass[letterpaper]{article} 
\usepackage[]{2026_arxiv}  
\usepackage{times}  
\usepackage{helvet}  
\usepackage{courier}  
\usepackage[hyphens]{url}  
\usepackage{graphicx} 
\usepackage{natbib}  
\usepackage{caption} 
\usepackage{algorithm}
\usepackage{algorithmic}
\usepackage{multirow}
\usepackage{booktabs}
\usepackage{xcolor} 

\usepackage{amssymb}       
\usepackage{array}          
\usepackage{makecell}       
\usepackage{bbding}        
\usepackage{amsmath}

\usepackage{newfloat}
\usepackage{listings}
\DeclareCaptionStyle{ruled}{labelfont=normalfont,labelsep=colon,strut=off} 
\floatstyle{ruled}
\newfloat{listing}{tb}{lst}{}
\floatname{listing}{Listing}
\newcommand{\methodname}{OmniEvent}

\title{\methodname: Unified Event Representation Learning}
\author {
    Weiqi Yan\textsuperscript{\rm 1}\footnote{Equal contribution.},
    Chenlu Lin\textsuperscript{\rm 1}\footnotemark[1],
    Youbiao Wang\textsuperscript{\rm 1},
    Zhipeng Cai\textsuperscript{\rm 2}\footnote{Corresponding author.},
    Xiuhong Lin\textsuperscript{\rm 1},
    Yangyang Shi\textsuperscript{\rm 2},
    Weiquan Liu\textsuperscript{\rm 3},
    Yu Zang\textsuperscript{\rm 1}\footnotemark[2]
}
\affiliations {
    \textsuperscript{\rm 1}Fujian Key Lab of Sensing and Computing for Smart Cities, School of Informatics, Xiamen University (XMU), China\\
    \textsuperscript{\rm 2}Meta AI\\
    \textsuperscript{\rm 3}College of Computer Engineering, Jimei University, Xiamen, China\\
    \{yanweiqi199888, czptc2h, shiyang1983\}@gmail.com, \{36920231153214, 36920241153251, xhlinxm\}@stu.xmu.edu.cn, wqliu@jmu.edu.cn, zangyu7@126.com
}

\usepackage{bibentry}

\begin{document}

\maketitle

\begin{abstract}
\label{sec:abstract}
Event cameras have gained increasing popularity in computer vision due to their ultra-high dynamic range and temporal resolution. However, event networks heavily rely on task-specific designs due to the unstructured data distribution and spatial-temporal (S-T) inhomogeneity, making it hard to reuse existing architectures for new tasks. 
We propose \methodname, the first unified event representation learning framework that achieves SOTA performance across diverse tasks, fully removing the need of task-specific designs. 
Unlike previous methods that treat event data as 3D point clouds with manually tuned S-T scaling weights, \methodname\ proposes a decouple-enhance-fuse paradigm, where the local feature aggregation and enhancement is done independently on the spatial and temporal domains to avoid inhomogeneity issues. Space-filling curves are applied to enable large receptive fields while improving memory and compute efficiency. The features from individual domains are then fused by attention to learn S-T interactions. The output of \methodname\ is a grid-shaped tensor, which enables standard vision models to process event data without architecture change. 
With a unified framework and similar hyper-parameters, \methodname\ out-performs (tasks-specific) SOTA by up to 68.2\% across 3 representative tasks and 10 datasets (Fig.~\ref{fig:teaser}).
Code will be ready in https://github.com/Wickyan/OmniEvent .

\begin{figure}[t]
\centering
\includegraphics[width = 0.53\textwidth]{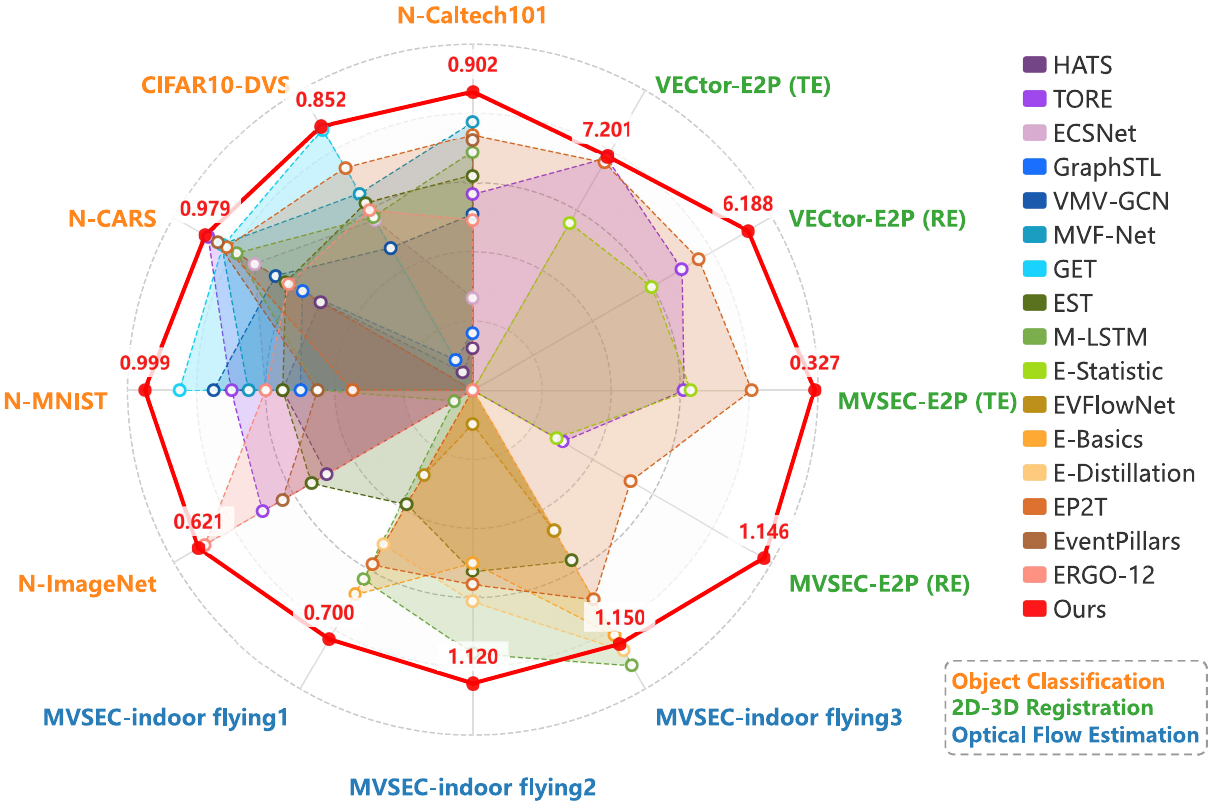}
   \caption{\textbf{Teaser.} We propose \methodname, the first unified event representation frameowork that achieves SOTA performance across diverse tasks (colors in the bottom right corner represent different tasks) without complex task-specific design or hyper-parameter tuning. (RE) and (TE) are rotation and translation errors.}
\label{fig:teaser}
\end{figure}

\end{abstract}

\section{Introduction}
\label{sec:intro}

Event cameras are a new type of cameras that only respond to pixel positions where the illumination change exceeds a certain threshold.
Compared with conventional cameras, event cameras have very high dynamic range (HDR, up to 120 dB~\cite{Event_survey}) and temporal resolution, very low response latency and power consumption. These advantages makes event cameras less susceptible to challenging illumination and motion. 
However, the output of an event camera consists of a stream of events that only encode the time, location and polarity (direction) of brightness changes. Consequently, each event alone carries very little information, and the event generation rate varies significantly in different scenarios. 
Furthermore, the unstructured data stream of event cameras is sparse in the spatial plane but may be dense (up to 300 million events per second~\cite{Meps_events}) in the temporal dimension. 
All of these challenges have spurred the community to develop specialized algorithms tailored to the unique data characteristics of event cameras.

Data-driven approaches have gained significant attention in recent years.
Given that event cameras are fundamentally 2D visual sensors, an intuitive approach is to leverage existing research on  frame-based convolutional neural networks (CNNs). To accomplish this, asynchronous event data is typically transformed into a grid-shaped feature tensor representation that can be synchronously updated.
Existing approaches predominantly rely on hand-crafted event tensorization methods, such as statistical features~\cite{Statistical_pn,Statistical_counting,Statistical_time,EV_FlowNet}, exponential~\cite{decays_kernel_HOTS,decays_kernel_HATS} or linear~\cite{decays_kernel_linear} decays, event surfaces~\cite{surface_Distance,surface_TORE}, and event voxel~\cite{voxel_optflow,voxel_imgrebuilt}. Constrained by their representational capacity, these hand-crafted event representation methods often lose S-T details and cannot be optimized for different downstream tasks.

Recent approaches in event-based representation learning have emerged by treating event streams as 3D point clouds. However, classical frameworks~\cite{pointnet++,EventNet} face dual challenges when applied to event data. First, Euclidean distance-based neighborhood construction struggles to adapt to spatiotemporal heterogeneity—fixed spatial-radius neighborhood searches prove inefficient for dynamically sparse event distributions, while microsecond-level timestamp differences are nearly obliterated in Euclidean distance calculations. Second, although the K-Nearest Neighbors (KNN) algorithm enables local feature aggregation, its receptive field remains constrained by local neighborhood size. Increasing the neighbor count \( K \) expands coverage but incurs heavy compute and memory costs, failing to fundamentally resolve distance degradation in high-dimensional spaces. Existing improvements, such as E2PNet~\cite{E2PNet}, attempt to mitigate these issues through spatiotemporal feature-weighted fusion. However, static weighting mechanisms cannot adapt to dynamic requirements across diverse scenarios.

We propose \emph{\methodname}, the \emph{first} unified representation learning framework effective for different event tasks. \methodname\ first decouples event data into spatial and temporal feature subspaces, where independent encoders capture high-precision geometric features and microsecond-level motion patterns. Subsequently, space-filling curve-guided feature aggregation maps 3D spatial coordinates into 1D continuous curve sequences. This strategy naturally establishes long-range spatial correlations through curve traversal, avoiding KNN computational bottlenecks while enabling the model to adaptively adjust fusion weights between local and global features. Combining with standard architectures~\cite{ResNet34,EV_FlowNet,LCD}, \methodname\ surpasses state-of-the-art methods across 3 event tasks and 10 datasets, advancing SOTA by a large margin (up to 68\% error reduction).
Our main \emph{contributions} are summarized below:
\begin{itemize}

\item We propose \methodname, the \emph{first} framework that can learn effective representations for various event vision tasks with a unified architecture.

\item We introduce the decouple-enhance-fuse paradigm, which \emph{fully} decouples spatial and temporal domains during early feature aggregation, eliminating their interference without task/dataset-specific scaling weight tuning.

\item We apply space filling curves for event representation learning which significantly enlarge the receptive fields and removes the compute and memory bottlenecks from KNN-based methods.

\end{itemize}

\section{Related Work}
\label{related_Works}

\paragraph{Hand-crafted Representations.}
Direct projections mark event pixel positions but ignore temporal information~\cite{Statistical_pn}. 
Statistical features~(event counts/timestamps)~\cite{EV_FlowNet,Statistical_counting,Statistical_time} and voxels~\cite{voxel_optflow,voxel_imgrebuilt} approximate spatiotemporal distributions but lack fine-grained correlations. 
Time decay representations~(exponential/linear)~\cite{decays_kernel_HOTS,decays_kernel_linear} and event surfaces~\cite{optflow_plane_fit,surface_Distance} preserve details through SNN-inspired structures~\cite{SNN_2019}, with improved versions using event buffering~\cite{surface_TORE} or neighborhood merging~\cite{decays_kernel_HATS,surface_neighbor} for stability. 
Graph-based methods~\cite{ST_graph_cuts,Graph_2020,AEGNN} excel spatially but lose temporal details with high computational costs. 
Overall, these representations fundamentally limit comprehensive spatiotemporal capture, hindering task adaptability.

\paragraph{Learned Representations.} 
SNNs directly process events but require specialized hardware~\cite{SNN_speHardware_optflow,SNN_speHardware_depth} and face training challenges~\cite{SNN_Gradient_descent,SNN_Conversion}. 
EST/DDES~\cite{EST_LR,FIFO_volume_learning} transform events via MLP-learned temporal kernels but remain representationally limited. 
Matrix-LSTM~\cite{Matrix_LSTM} encodes patch sequences but handles only short-range dependencies with custom parallelization needs. 
ECSNet~\cite{ECSNet} and E2PNet process events as point clouds but suffer from Euclidean metric constraints and inefficient long-range modeling. 
\methodname~overcomes these by spatiotemporally decoupling dimensions, improving efficiency and correlation capture.

\section{Method}
\label{method}

\begin{figure*}
\centering
\includegraphics[width = 0.93\textwidth]{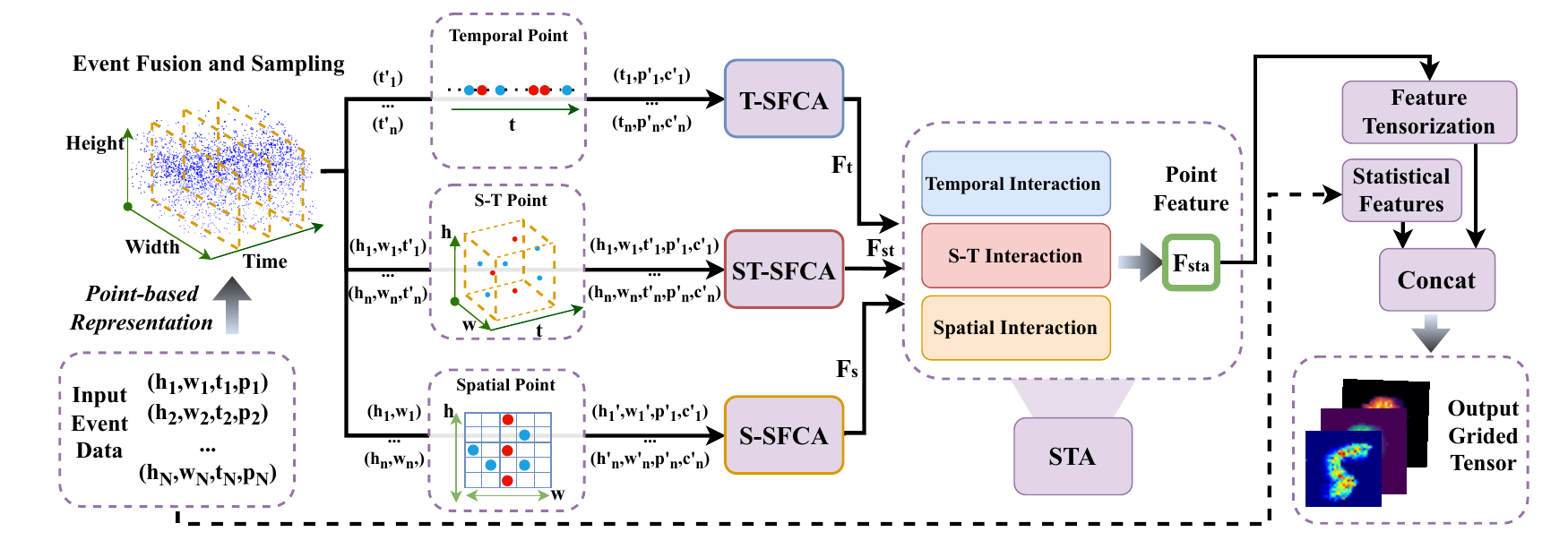}
   \caption{\textbf{\methodname\ architecture.} \methodname\ treats event data as spatio-temporal point clouds. Event Fusion and Sampling (EFS) fuses and samples unstructured event data into a fixed number of points.
   Three sets of Space-Filling-Curve-based Aggregation (SFCA) modules independently extract spatial-only, temporal-only and spatial-temporal information using different distance metrics and features, effectively addressing spatio-temporal inhomogeneity.
   Subsequently, Spatio-temporal Separated Attention (STA) models the spatio-temporal correlations of features using attention. 
   Finally, Feature Tensorization (FT) converts unstructured event features into a grid-shaped tensor and combines them with event statistical features. The output grid-shaped tensor can be used by any frame-based method. 
   }
\label{fig:Overview}
\end{figure*}


During event data collection, each pixel of the event camera responds independently and asynchronously to illumination changes. 
Specifically, given an event camera of size $(H, W)$, and $I(t_i, h, w)$ as the illumination at time $t_i$ for the pixel at $(h, w)$. When the inequality
\begin{equation}
 \left\|\log (I(t_i, h, w))-\log(I(t_{i-1}, h, w))\right\|> \tau
 \label{eq1}
\end{equation}
is satisfied, this pixel will generate an \textit{event}, which is expressed as $\mathbf{e}=(h,w,t_i,p)$. The polarity $p\in\{+1,-1\} $ indicates the brightness change direction (increasing or decreasing) and $\tau$ is the trigger threshold. $t_{i-1}$ represents the time when the event was triggered last time at pixel $(h,w)$. 
 
\methodname\ treats event data as spatio-temporal point clouds. 
As the imaging plane $(H, W)$ and the time axis can be combined to establish a three-dimensional spatio-temporal coordinate system $(h, w, t_i)$. 
Such a representation preserves the original information and data sparsity. 

As illustrated in Fig.~\ref{fig:Overview}, \methodname\ consists of five consecutive modules: Event Fusion and Sampling (EFS), Spatio-Temporal Decoupling (STD), Space Filling Curve based Aggregation (SFCA), Spatio-temporal Separated Attention (STA), and Feature Tensorization (FT).

The framework design follows a \emph{decouple}-\textbf{enhance}-\underline{fuse} paradigm. 
\emph{EFS} (Sec.~\ref{sec:Sampling}) fuses the event points and subsamples them into a fixed number to enable efficient computation and minimize information loss. 
\emph{STD} (Sec.~\ref{sec:STD}) proposes a radical decoupling of spatial and temporal dimensions. This approach reduces the dependence on joint distance metric optimization and enables the capture of long-range spatiotemporal correlations through separate temporal and spatial neighborhood definitions.
\textbf{SFCA} (Sec.~\ref{sec:aggregation}) conducts sequential feature enhancement on decoupled features by fully utilizing event presentation.
SFCA also decouples positional embeddings from feature aggregation and the event fusion sampling method, achieves SOTA representation capabilities while improving algorithm efficiency.
\underline{STA} (Sec.~\ref{sec:STA+FT}) reorganizes the decoupled temporal and spatial features into global features, modeling spatiotemporal correlations. 
\underline{FT} (Sec.~\ref{sec:STA+FT}) uses sparse convolution to convert unstructured spatiotemporal point features into grid-shaped tensors and introduces event statistical features to compensate for the global information loss due to sampling. FT enables \methodname\ to focus on extracting spatiotemporal correlations and detailed information. The output grid-shaped tensor can be integrated into any frame-based network, allowing it to process event data without changing the architecture/training process.




\subsection{Event Fusion and Sampling (EFS)}
\label{sec:Sampling}

Although learning architectures based on 3D points~\cite{pointnet++,PointCNN,PosPool} can adapt to irregular and discrete event point clouds, they are still limited by the number of points they can process simultaneously. 
Fortunately, we found significant information redundancy within events. For instance, moving objects with high contrast against the background may trigger numerous events simultaneously at the same pixel location (see eq.(\ref{eq1})). This redundancy allows us to perform event point fusion before sampling.


As illustrated in the top left of Fig. \ref{fig:Overview}, Similar to event voxelization~\cite{voxel_optflow,voxel_imgrebuilt}, we divide a batch of events into 
$T$ temporal segments, where events within the same pixel and temporal segment are fused. Specifically, given $N$ events $\mathbf{e_{hw}}=(h,w,t_i,p_i)$, $i\in\{1,...,c\} $ 
on position $(x, y)$ in time slice $T_n$, the fused event can be represented as $\mathbf{e'_{hw}}=(h, w, t_{avg}, p_{acc}, c)$, $t_{avg}=\frac{1}{c}\sum_{i} t_i$ and $p_{acc}=\sum_{i} p_i$. Inspired by eHPE~\cite{eHPE}, the number $c$ of events involved in the fusion is added as a new feature to the fused event point cloud. 
Unlike voxelization methods~\cite{voxel_optflow,voxel_imgrebuilt} that use the time coordinates corresponding to voxel blocks, we retain the precise average time and the number of events involved in fusion, minimizing information loss to the extent possible.
The number of temporal segments $T$ during event fusion can be balanced between temporal precision and sampling loss based on different tasks.

Random sampling is used to select top $M$ inputs from the fused event point cloud for later processing. Random sampling significantly reduces computational costs~\cite{pointnet++} during sampling and does not require additional virtual points~\cite{E2PNet} for feature propagation.
To reduce the interference caused by the differing numerical scales (pixels vs. seconds) in event point cloud computation, each input point $\mathbf{e'} = (h, w, t', p', c')$ is normalized as $\mathbf{\hat{e}}=(\frac{h}{W}, \frac{w}{W}, \frac{t'}{t_\text{max}-t_\text{min}}, p, c)$, where $t_\text{max}$ and $t_\text{min}$ are the maximum and minimum timestamps of the input point cloud. By normalizing the event feature scales, the aggregation stage helps to avoid feature imbalance due to scale differences, facilitating a more uniform extraction of event features across the spatiotemporal dimensions.

\subsection{Spatio-Temporal Decoupling (STD)}
\label{sec:STD}
Traditional point cloud networks often use Euclidean distance for neighbor indexing. However, the heavy spatiotemporal scale variation across different event data limits the applicability of Euclidean distance for feature aggregation. E2PNet introduces a spatiotemporal weighted distance to mitigate the dilution of local feature diversity caused by spatiotemporal heterogeneity. However, the need for task/data-specific weight tuning limits its generalization.

We argue that despite the imbalance in spatiotemporal scales, there still exists an inherent relationship between space and time. However, explicitly modeling this relationship often leads to coarse granularity and inefficiency. Therefore, we propose separating spatiotemporal information to obtain independent unimodal distance metrics, allowing the scale of neighboring points to adapt to the current modality and ensuring a balanced trade-off between sparsity and density. Subsequently, the local features extracted through the spatiotemporal decoupling feature extraction branch are further refined in the STA phase, where the network uses an attention module to fuse the spatiotemporal scaling relationship, thereby enabling diverse combinations of local features and enhancing their representational power.
Specifically, given a set of fused, sampled, and normalized events $\mathbf{\hat{e}}=(h, w, t', p', c')$ from EFS. As illustrated in Fig.~\ref{fig:Overview}, The sampled event point clouds are separately fed into temporal, spatial, and spatio-temporal SFCA modules. In S-SFCA, we solely use space as the distance metric to determine neighborhoods, conversely, T-SFCA uses time as the distance metric. ST-SFCA adopts the traditional Euclidean distance.
We keep ST-SFCA with the decoupled branches since there is a strong inherent relationship between space and time. The decoupled spatiotemporal local features aim to optimize the collaborative expression of these features. Later ablation study demonstrate the necessity and effectiveness of all three branches.

\subsection{Space Filling Curve based Aggregation (SFCA)}
\label{sec:aggregation}

\begin{figure}[t]
\centering
\includegraphics[width = 0.45\textwidth]{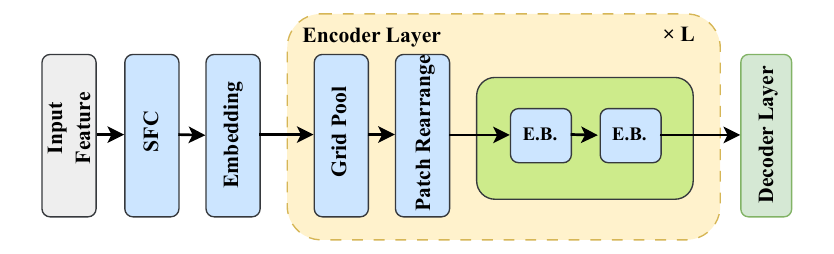}
\caption{\textbf{SFCA structure.} The input undergoes serialization using a space-filling curve (SFC), followed by an embedding layer. Subsequently, the data passes through L Encoder Layers (for S-SFCA and T-SFCA combinations, L=2; for ST-SFCA configurations, L=4) , each of which integrates Grid Pool operations with dynamic patch rearrangement for adaptive feature grouping. The processed features then flow through 2 consecutive encoder blocks (E.B.s).  The decoder layer finally generates the output.}
\label{fig:SFCA}
\end{figure}

Unlike conventional point clouds, event data requires large receptive fields to model multi-scale spatiotemporal correlations. While traditional methods derive semantics from precise boundary features, event streams encode semantic information through temporal-spatial evolution patterns. Previous methods are limited by the receptive fields: PointNet-based methods~\cite{PointNet,pointnet++} suffers from information discontinuity across sampling hierarchies and severe signal decay in deep layers; KNN-based methods~\cite{zhao2021point,wu2022point} face computational bottlenecks (linear complexity with neighbor counts) when scaling to dense event streams. 

Our work addresses this through a space-filling curve-driven module inspired by~\cite{pointTransformersV3}. This strategy enables efficient large-receptive-field learning by establishing continuous multi-scale correlations, directly aligning with the spatiotemporal nature of event data while maintaining computational feasibility.



Given a set of event points \( P = \{({s'}_i, {p'}_i, {c'}_i)|\ i = 1,2,...,N\} \), where \(s'\) is the distance metric dimension(s) of the current SFCA branch, we first map it to a regular grid structure by applying \(\left\lfloor {s_i'} / {g} \right\rfloor\) based on a given grid size \( g \). This allows us to use the inverse mapping of a space-filling curve, to transform each point in \(s'\) into a one-dimensional encoding value as follows:
\begin{equation}
    V_m = \phi_m\!\left(\left\lfloor\frac{s_i'}{g}\right\rfloor\right)
     \label{eq2}
\end{equation}
where \(\phi_m(\cdot)\) represents different types of space-filling curves (e.g., the Z-Order curve~\cite{morton1966computer} or the Hilbert curve~\cite{hilbert1935neubegrundung}) and \(V_m \in \mathbb{Z}\) denotes the 1-D index of point \(m\) on a space-filling curve. The event points \( P \) are then embedded into a feature representation \( F^0 \in \mathbb{R}^{N \times C^{0}} \), which is then fed into the encoder layer.

To enhance the model's generalization ability, in each encoder layer, we first randomly select a precomputed encoding value \( V_m \) from one of the space-filling curves and use it to sort the event points. 
The Hilbert curve is chosen as the encoding foundation due to its superior ability to preserve spatial continuity. By mapping 2D coordinates into a 1D sequence, it effectively maintains the spatial proximity of physically adjacent points. In contrast, the Z-order curve introduces spatial discontinuities due to its alternating coordinate encoding; however, this very property makes it well-suited for capturing non-contiguous event distributions.

In S-SFCA and T-SFCA, the expansion of the local receptive field does not rely on the jump-based connections of the Z-order curve. In fact, such discontinuous encoding could interfere with event feature extraction, which is why we exclusively adopt the Hilbert curve in these modules. However, in ST-SFCA, capturing temporal variations at the same pixel location proves challenging using the Hilbert curve alone, as its strong spatial continuity fails to account for long-range temporal dependencies. Here, the discontinuous connections of the Z-order curve become advantageous, allowing for the effective capture of local temporal variations across distant time steps. Therefore, we employ both the Hilbert and Z-order curves in our design.

Then, we partition \( F \) into \( \left\lfloor N / p \right\rfloor \) non-overlapping patches \( \hat{F}_j \) \( \in \mathbb{R}^{p \times C^{0}}\) (\( j = 1, 2, \dots, \left\lfloor N / p \right\rfloor \)), where \( p = 512 \) in our model, with each patch serving as a local neighborhood for feature extraction. This partitioning is achieved via encoding-based sorting with a linear time complexity of \( O(N) \). In contrast, traditional KNN-based neighborhood search has an \( O(NK) \) time/memory complexity where $K$ is the number of neighboring points, e.g., 512 in our case. The immense computational burden makes it infeasible to significantly expand the receptive field, limiting previous methods to use only 16 points~\cite{E2PNet}.

Each patch undergoes two self-attention encoder blocks. The first block transforms input patch features \( \hat{F}_j \) into intermediate representations \( \hat{F}^{1}_j \), while the second block refines them into \( \hat{F}^{2}_j \in \mathbb{R}^{P \times C^1} \), where \( C^1 \) is the output channel size of the first encoder layer.

Subsequently, the encoded features \( F^1 \) are fed into the next encoder layer. After selecting the \( V_m\), we introduce a pooling operation before sorting. Specifically, we perform a right shift by \( y \) bits on the spatial encoding values of \( V_m \), ensuring that originally adjacent points are mapped to the same value:
\begin{equation}
V'_m = V_m \gg y
\end{equation}
When selecting \( y \), unlike the fixed-value approach in~\cite{pointTransformersV3}, we account for the unique characteristics of event cameras. Events tend to be spatially dense but temporally sparse. 
This imbalance necessitates dynamic downsampling: we set \( y = 5 \) initially to reduce redundancy in densely clustered spatial points while preserving key structural features, then adjust it to \( y = 3 \) in later stages to balance the decreasing event density over time while maintaining a uniform distribution.

Based on this, we perform max pooling on points with the same \( V'_m \) value to achieve aggregation. This reduces the point count from \( N \) to \( N^1 \approx N / 2^y \), generating downsampled features \( F^1_d \in \mathbb{R}^{N^1 \times C^1} \). Then those downsampled points are processed similarly to the first layer, being fed into the encoder block for transformation.

After L layers hierarchical stacking, the model achieves exponential receptive field growth. The base coverage \( P \) defines the initial local perception, while \( 2^{yL} \) represents the cumulative expansion factor from \( L \) pooling layers. The total receptive field after \( L \) layers is:
\begin{equation}
R_L = P \cdot 2^{yL}
\end{equation}
Theoretically, the final receptive field of SFCA would be 32 times bigger than K-NN based methods E2PNet.



\begin{figure}[t]
\label{attention_structure}
\centering
\includegraphics[width = 0.5\textwidth]{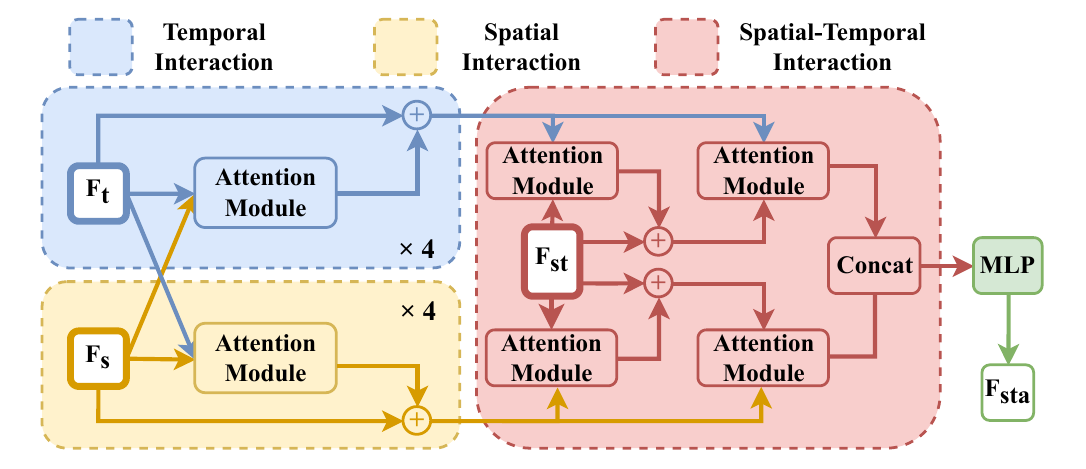}
   \caption{\textbf{STA structure.} STA integrates the temporal, spatial, and spatiotemporal features that were previously decoupled and encoded, completing the \emph{decouple-enhance-fuse} paradigm. It enhances the temporal-only and spatial-only features from SFCA through four rounds of mutual attentions, and then refines these features by conducting multiple rounds of attentions with the spatiotemporal features. The refined features are fed into a multi-layer perceptron (MLP), yielding the output feature $F_{sta}$.}
\label{fig:attention}
\end{figure}
\subsection{Spatio-temporal Separated Attention (STA) and Feature Tensorization (FT)}
\label{sec:STA+FT}

After completing feature aggregation, to further establish the correlation between time and space, we feed the features from the three SFCA branches into STA for feature computation based on attention mechanisms. The structure of STA is illustrated in Fig.~\ref{fig:attention}, see Appendix~\ref{appdx:Details_of_STA} for Attention Module architecture details.

The features \( F_s \) and \( F_t \) pass through the spatiotemporal cross-attention fusion module to capture the relationships between decoupled spatial temporal dimensions. The enhanced features obtained from this process are then added residually to the input, forming \( F_s^1 \) and \( F_t^1 \) as the input for the next interaction stage. After four stages of spatiotemporal cross-attention fusion, \( F_s^4 \) and \( F_t^4 \) are globally fused with \( F_{st} \) in the Spatial-Temporal Interaction module. In this step, \( F_s^4 \) and \( F_t^4 \) serve as representative features in the spatial and temporal distance measurement domains to enhance the global feature \( F_{st} \). Finally, the bidirectional enhancement results are concatenated, and the concatenated features are mapped through an MLP to obtain the output \( F_{sta} \).


\section{Experiments}
\label{sec:experiments}

To validate the general applicability of \methodname, we conduct experiments on 3 representative and diverse tasks: classification(Sec.~\ref{sec:Classification}), optical flow estimation(Sec.~\ref{sec:Optical_Flow}) and event to point cloud Registration(Sec.~\ref{sec:Event-to-Point_Cloud_Registration}).
Sec.~\ref{sec:Ablation} further performs detailed analysis, validating the effectiveness of individual components.

\subsection{Classification}
\label{sec:Classification}

\paragraph{Datasets.} 
We conduct experiments on four event-based object classification datasets: N-Caltech101~\cite{Caltech101_MINIST}, N-MNIST~\cite{Caltech101_MINIST}, CIFAR10-DVS~\cite{Cifar10}, N-CARS~\cite{decays_kernel_HATS}, and N-ImageNet~\cite{kim2021n}. 
The datasets N-Caltech101, N-MNIST, CIFAR10-DVS, and N-ImageNet are generated by recording moving images on an LCD screen using an event camera. 
Following the common practice, we keep 80\% of the data for training and validation, and the remaining 20\% for testing. We use the N-Caltech101 partition from~\cite{EST_LR, Matrix_LSTM}.

\begin{table}[h]
\caption{\textbf{Classification result}. We report top-1 classification accuracy, which is the best values from the original papers and our reproduced results. \textbf{Bold} and \underline{underlined} values represent the best and second-best results. The upper part of the table shows SOTA task-specific methods, including HATS~\cite{decays_kernel_HATS}, TORE~\cite{surface_TORE}, ECSNet, GraphSTL~\cite{GraphSTL}, VMV-GCN~\cite{VMV-GCN}, MVF-Net~\cite{MVF-Net} and GET~\cite{GET}      
.
The lower part shows general event representation learning methods with ResNet34, including EST, M-LSTM~\cite{Matrix_LSTM} and E2PNet, EventPillars~\cite{fan2025eventpillars} and ERGO-12~\cite{zubic2023chaos}}

\label{tab:classification}
\centering
\fontsize{7.5pt}{6.5pt}\selectfont  
\setlength{\tabcolsep}{1mm}  

\begin{tabular}{@{}p{1.8cm} p{1.0cm} p{1.5cm} p{1.1cm} p{1.1cm} p{1.5cm}@{}}
\toprule
Method & CIFAR10-DVS & N-Caltech101 & N-CARS & N-MNIST & N-ImageNet \\ \midrule
HATS & 52.4 & 64.2 & 90.2 & 99.1 & 47.1 \\
TORE & N/A & 79.8 & \underline{97.7} & 99.4 & 54.6 \\  
ECSNet & 72.7 & 69.3 & 94.6 & 99.2 & N/A \\
GraphSTL & 54.0 & 65.7 & 91.4 & 99.0 & N/A \\
VMV-GCN & 69.0 & 77.8 & 93.2 & 99.5 & N/A \\
MVF-Net & 76.2 & \underline{87.1} & 96.8 & 99.3 & N/A \\
GET & \underline{84.8} & N/A & 96.7 & \underline{99.7} & N/A \\ 
\midrule
\multicolumn{6}{c}{General Event Representation Learning (ResNet34)} \\ 
\midrule
EST & 74.9 & 81.7 & 92.5 & 99.1 & 48.9 \\  
M-LSTM & 73.1 & 84.1 & 95.8 & 98.9 & 32.2 \\ 
E2PNet & 79.6 & 85.8 & 96.5 & 98.7 & N/A \\
EventPillars & N/A & 85.3 & 97.1 & 98.9 & 52.3 \\
ERGO-12 & 74.0 & 77.2 & 92.3 & 99.2 & \underline{61.4} \\
\methodname(Ours) & \textbf{85.2} & \textbf{90.2} & \textbf{97.9} & \textbf{99.9} & \textbf{62.1} \\ 
\bottomrule
\end{tabular}
\end{table}

\paragraph{Implementation.} 

Event representation learning focuses on feature extraction and representation transformation, so evaluating its performance requires selecting an appropriate task network. To ensure a fair comparison, we follow the settings of existing representation learning methods~\cite{EST_LR} and use a basic ResNet34 as the task network.
Our SFCA (Fig.~\ref{fig:SFCA}) consists of multiple layers of Sub-SFCA. Specifically, T-SFCA and S-SFCA each contain 2 layers, while ST-SFCA comprises 4 layers. Detailed architectural information for each layer can be found in the appendix. We follow EST~\cite{EST_LR} to use ADAM optimizer for all experiments with a learning rate of $10^{-4}$. Due to memory constraints, we set the batch size to 10. 
We perform early stopping on a validation set in all experiments.
All methods are implemented using PyTorch on an Intel Xeon Gold 6342 CPU and an NVIDIA A40 GPU.

As shown in Tab.~\ref{tab:classification}, \methodname\ outperforms both task-specific and general event based methods. It is worth noting that \methodname, paired with a basic task network and devoid of any training hyperparameter tuning, achieves leading performance on \emph{all} datasets. Also \methodname is the \emph{only} method that is consistently better than others, the second best method varies across datasets, indicating limited generalization of existing frameworks.

\subsection{Optical Flow Estimation}
\label{sec:Optical_Flow}

\begin{table*}[ht]
\caption{\textbf{Optical flow estimation result on MVSEC. }
AEE represents the average end-point error and Out. represents the proportion of AEE errors exceeding 3 pixels. The upper part of the table presents task-specific methods, the bottom part compares representation learning methods combined with the EVFlowNet~\cite{EV_FlowNet}. \textbf{Bold} and \underline{underlined} values represent the best and second-best results.}
\label{tab:Opt_Flow}
\centering

\fontsize{9pt}{3pt}\selectfont
\setlength{\tabcolsep}{3mm}  

\begin{tabular}{@{}lllllll@{}}
\toprule
\multicolumn{1}{c}{\multirow{2}{*}{Methods}} & \multicolumn{2}{c}{indoor flying1} & \multicolumn{2}{c}{indoor flying2} & \multicolumn{2}{c}{indoor flying3} \\ 
\cmidrule(lr){2-3} \cmidrule(lr){4-5} \cmidrule(lr){6-7}
& AEE(↓) & \%Out.(↓) & AEE(↓) & \%Out.(↓) & AEE(↓) & \%Out.(↓) \\ \midrule
EVFlowNet(EVF.) & 1.03 & 2.20 & 1.72 & 15.10 & 1.53 & 11.90 \\
E-Basics~\cite{E-Basics} & \underline{0.79} & 1.20 & 1.40 & 10.90 & 1.18 & 7.40 \\
E-Distillation~\cite{E-Distillation} & 0.89 & 0.66 & 1.31 & 6.44 & \underline{1.13} & \underline{3.53} \\ \midrule
EST+EVF. & 0.97 & 0.91 & 1.38 & 8.20 & 1.43 & 6.47 \\
M-LSTM+EVF. & 0.82 & 0.53 & \underline{1.19} & \textbf{5.59} & \textbf{1.08} & 4.81 \\
E2PNet+EVF. & 0.85 & \underline{0.05} & 1.35 & 6.75 & 1.30 & 4.09 \\
\methodname(Ours)+EVF. & \textbf{0.70} & \textbf{0.03} & \textbf{1.12} & \underline{6.36} & 1.15 & \textbf{3.33} \\ \bottomrule
\end{tabular}

\end{table*}

\paragraph{Datasets.} 
We use the widely recognized Multi Vehicle Stereo Event Camera Dataset (MVSEC)~\cite{MVSEC} dataset for optical flow experiments. MVSEC contains indoor and outdoor scene sequences with significant differences and provides optical flow ground-truth. 


\paragraph{Implementation.} 

Following previous works~\cite{EST_LR, Matrix_LSTM}, we use the classic EV-FlowNet~\cite{EV_FlowNet}  (simple U-net) as the task network, training on the \textit{outdoor\_day1}, and \textit{outdoor\_day2} sequences with optical flow ground truth as supervision. 
Following~\cite{EST_LR}, we evaluate models by comparing the average end-point error (AEE = $\frac{1}{N} \sum_i |f_{est} - f_{gt} |_2 $) and global outlier (where AEE $>3$) ratio  on the indoor flying sequences, which have different scenes and motion patterns from the training set to validate generalization. We also mask the hood area to avoid interference from reflections following E-cGAN~\cite{E-cGAN}.
Additionally, due to the sparsity of events, we evaluate performance only at the pixel locations where events occur (sparse optical flow), as done in~\cite{EST_LR, EV_FlowNet}.
Due to memory limitations, the batch size is set to 6 in optical flow estimation. 
Apart from the batch size, \methodname's structure and other training parameters \emph{remain the same} as in the classification tasks (Sec.~\ref{sec:Classification}). 


As shown in Tab.~\ref{tab:Opt_Flow}, \methodname\ also surpasses all other representation learning methods. 
See Appendix~\ref{appdx:Visualization_flow} for visualization of optical flow estimation.
It is worth noting that these specialized optical flow estimation networks incorporate additional information. For example, E-Distillation employs knowledge distillation from optical flow estimation networks based on traditional images. 
Despite using a simple U-Net as the task architecture \emph{without hyper-parameter tuning}, \methodname\ remains competitive against specialized optical flow estimation networks.

\subsection{Event-to-Point Cloud Registration}
\label{sec:Event-to-Point_Cloud_Registration}

\begin{figure}[t]
\centering
\includegraphics[width = 0.45\textwidth]{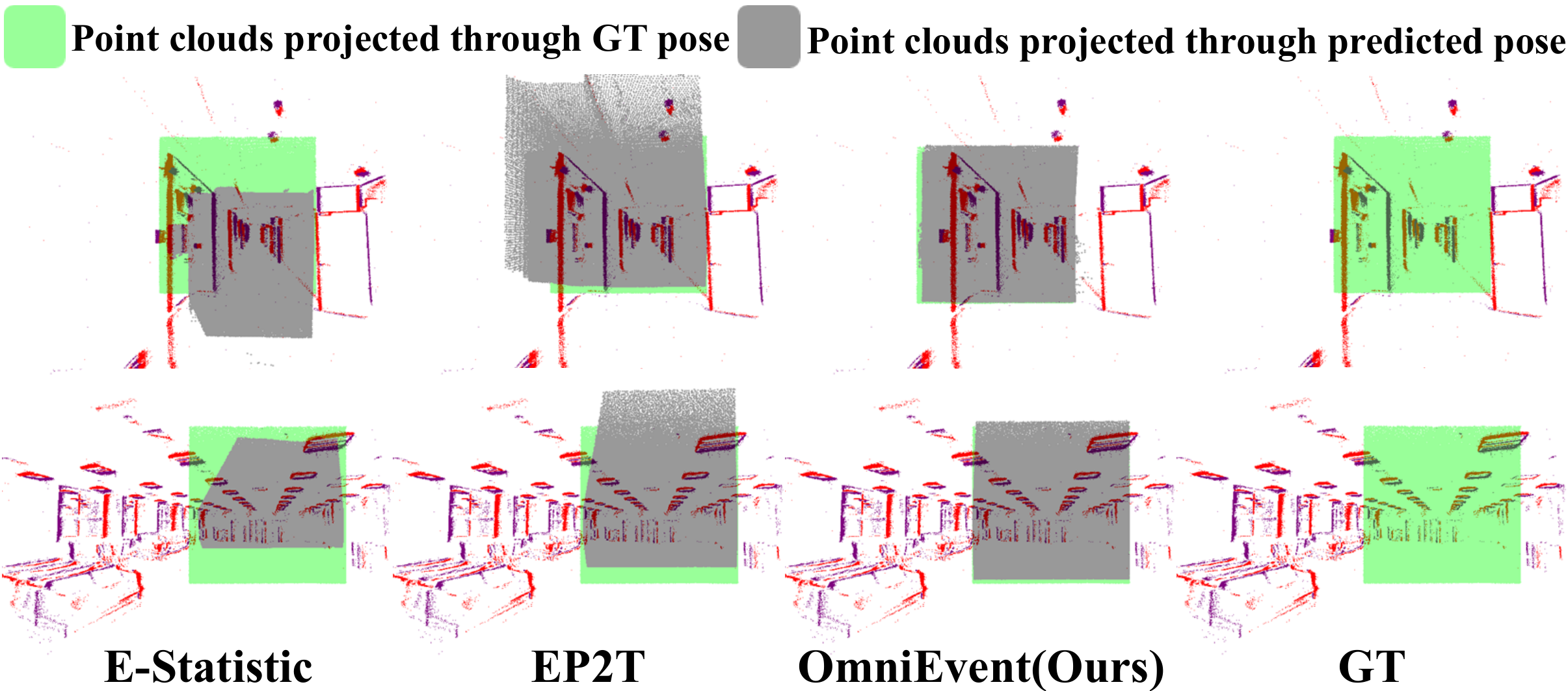}
   \caption{\textbf{Qualitative comparisons.} The \textit{green} box and the \textit{gray} box represent the projection on the 2D view after the 3D point cloud is rigidly transformed to the camera coordinate system using the true pose and the predicted pose. 
   } 
\label{fig:2d_3d_experiments}
\end{figure}

\paragraph{Datasets.}
Following the previous SOTA~\cite{E2PNet} we evaluate the performance using MVSEC and VECtor~\cite{VECtor} datasets. The train-test split follows E2PNet. 


\paragraph{Implementation.}
Following~\cite{EST_LR,E2PNet}, we use the classic LCD~\cite{LCD} as the task networks. 
Consistent with E2PNet, we evaluate models using the translation error $TE=\left\|T_{GT}-T_{pred}\right\|_{2}$ and the rotation error $RE=\arccos\left(\frac{\mathrm{tr}\left(R_{GT}^{-1}R_{pred}\right)-1}{2}\right)$. Here, T and R denote the translation vector and rotation matrix. The notation $\text{tr}(\cdot)$ represents the trace of a matrix. The subscripts “GT” and “pred” indicate the ground-truth and predicted poses, respectively. The translation error TE is measured in meters, while the rotation error RE is expressed in degrees. We use the same hyper-parameters as in the classification task except that the batch size is set to 4.

\begin{table}[t]
\scriptsize
\renewcommand{\arraystretch}{1.0}
\setlength{\tabcolsep}{1pt}
\centering
\caption{\textbf{2D-3D registration result.} Since the 3 handcrafted methods~\cite{EV_FlowNet,zhu2018unsupervisedeventbasedlearningoptical,mostafavi2021learning} are mutually enhancing, we combine them to optimize performance, and call the combined method E-Statistic. Additionally, we include two other methods, TORE and E2PNet, for a comprehensive evaluation. \textbf{Bold} and \underline{underlined} values represent the best and second-best results.}

\label{tab:E2P_Registration}
\setlength{\tabcolsep}{1pt}
\fontsize{8pt}{8pt}\selectfont

\begin{tabular}{@{}llcccc@{}}
\toprule
\multirow{2}{*}{Input} & \multirow{2}{*}{Method} & \multicolumn{2}{c}{MVSEC-E2P} & \multicolumn{2}{c}{VECtor-E2P} \\ 
\cmidrule(lr){3-4} \cmidrule(lr){5-6}
& & RE(°)(↓) & TE(m)(↓) & RE(°)(↓) & TE(m)(↓) \\ \midrule
Traditional Image & Grayscale Image & 6.335 & 1.347 & 17.879 & 13.200 \\ \midrule

\multirow{4}{*}{Event(Tensor-based)}  
& E-Statistic
& 4.968 & 1.297 & 11.034 & 9.416 \\
& TORE & 4.855 & 1.350 & {9.521} & \underline{7.254} \\
& E2PNet & \underline{3.606} & \underline{0.821} & \underline{8.672} & 7.403 \\
& \methodname(Ours) & \textbf{1.146} & \textbf{0.327} & \textbf{6.188} & \textbf{7.201} \\ \bottomrule
\end{tabular}
\end{table}

As shown in Tab.~\ref{tab:E2P_Registration}, \methodname\ outperforms previous SOTA by a large margin on both datasets. Compared with the best previous model E2PNet, \methodname\ reduces the RE and TE by up to $68\%$ and $60\%$ respectively. Fig.~\ref{fig:2d_3d_experiments} further visualizes the results of different methods. \methodname\ provides more accurate reprojections over SOTA methods.


\subsection{Analysis}
\label{sec:Ablation}

\begin{table}[t]
\caption{\textbf{Ablation}. Removing modules from \methodname\ hurt the classification accuracy in CIFAR10-DVS.}
\label{tab: ablation}
\centering

\renewcommand{\arraystretch}{1.2}

\setlength{\tabcolsep}{1pt}
\fontsize{9pt}{8pt}\selectfont

\begin{tabular}{ccccccc}
\hline
No. & S-SFCA     & ST-SFCA    & T-SFCA     & STA       & \makecell{Statistical \\ Features} & \makecell{Accuracy (\%)} \\\hline
1   & \checkmark & \checkmark & \checkmark & \checkmark & \checkmark & 85.2 \\
2   &            & \checkmark & \checkmark & \checkmark & \checkmark & 75.4 \\
3   & \checkmark &            & \checkmark & \checkmark & \checkmark & 79.2 \\
4   & \checkmark & \checkmark &            & \checkmark & \checkmark & 74.7 \\
5   & \checkmark & \checkmark & \checkmark &            & \checkmark & 76.7 \\
6   & \checkmark & \checkmark & \checkmark & \checkmark &            & 82.0 \\ \hline
\end{tabular}

\end{table}

\paragraph{Ablation.} To validate the effectiveness of different modules in \methodname, we conduct ablation studies by removing individual components from \methodname and compare the performance difference. For SFCA, we simply remove the corresponding branches. For STA, we replace it by feature concatenation from 3 SFCA branches. As shown in Tab.~\ref{tab: ablation}, the classification accuracy reduces as components are removed, proving the importance of each component. Tab.~\ref{tab: positional_encoding}  in Appendix~\ref{appdx:neighborhood_size} further shows the importance of large neighborhood size in SFCA. Increasing the neighborhood size from 16 to 512 points significantly enhances the performance.


Compared with the previous best model E2PNet, one advantage of \methodname\ is the removal of manually tuned weights for different spatial-temporal branches, which greatly simplifies the model training on different tasks. 
As shown in Fig.~\ref{fig:weights} of Appendix~\ref{appdx:weight_configurations},  E2PNet requires different weight combinations to achieve optimal performance for different tasks/datasets. And even though heavy weight tuning is conducted, E2PNet still performs worse than \methodname\ for all datasets.



\begin{table}[]
\caption{\textbf{Time and memory efficiency evaluated on CIFAR10-DVS.} \textbf{Bold} and \underline{underlined} values represent the best and second-best results.}
\label{tab: Effectiveness_efficiency}
\centering
\renewcommand{\arraystretch}{1.2}
\setlength{\tabcolsep}{5pt} 
\fontsize{9pt}{7pt}\selectfont

\begin{tabular}{@{}lcccc@{}} 
\toprule
Method       & \methodname (Ours) & EP2T & ECSNet & EST \\ \midrule
Time (ms)    & \underline{36.6} & 42.8 & 67.8  & \textbf{6.3} \\
Space (MB)   & 3146    & 4650 & \underline{2530} & \textbf{2457}  \\
Accuracy (\%)& \textbf{85.2}    & \underline{79.6} & 72.7 & 74.9 \\ \bottomrule
\end{tabular}

\end{table}

\paragraph{Memory and Time Efficiency.}
Besides the SOTA accuracy, \methodname\ also provides reasonable efficiency. As shown in Tab.~\ref{tab: Effectiveness_efficiency}, while EST achieves the fastest inference time, it sacrifices classification accuracy heavily, falling 10.3\% behind our method. \methodname\ attains the highest accuracy (85.2\%) with a competitive run time (36.6 ms), demonstrating a 14.4\% speed improvement over EP2T and 45.9\% faster than ECSNet. Meanwhile, \methodname\ requires only 3146MB memory, which is $\frac{1}{3}$ lower than EP2T. 
\section{Conclusion}
\label{sec:conclusion}
We propose \methodname, the \emph{first} framework that can learn effective representations across various event camera tasks with the \emph{same} architecture.
\methodname\ proposes a novel decouple-enhance-fuse paradigm, which fully decouples spatial-temporal domains during local feature aggregation to handle spatial-temporal inhomogeneity, and learn global spatial-temporal relationships during the  ``fuse" stage with attention mechanisms. 
\emph{Without} task-specific architecture design or hyper-parameter tuning, \methodname\ advanced SOTA across 3 representative tasks and 10 datasets. While \methodname\ provides a unified architecture, our current experiments still train separate models for different tasks due to the limited dataset size. An interesting future direction is to scale up model training and produce a single foundation model for various tasks.

\bibliography{2026_arxiv}

\clearpage
\appendix

\section{Appendix——\methodname: Unified Event Representation Learning}
\label{appdx:appendix}

\subsection{Details of SFCA Module}

The detailed configurations and parameters of the SFCA module are comprehensively presented in Tab.~\ref{tab:appendix_SFCA_settings1}, ~\ref{tab:appendix_SFCA_settings2}.

\label{appdx:Details_of_SFCA}

\begin{table}[h!]

    \centering
    \caption{\textbf{S-SFCA/T-SFCA model settings.}} 
    \label{tab:appendix_SFCA_settings1}
    
    \begin{tabular}{lp{40mm}}
    \toprule
    Config & Value \\  
    \midrule
    input channels & 5 \\
    order & ``hilbert", ``hilbert-trans" \\
    stride & [2, 2] \\
    encoder depths & [2, 2, 2] \\
    encoder channels & [64, 128, 256] \\
    encoder num head & [4, 8, 16] \\
    encoder patch size & [512, 512, 512] \\
    decoder depths & [2, 2] \\
    decoder channels & [64, 128] \\
    decoder num head & [4, 8] \\
    decoder patch size & [512, 512] \\
    mlp ratio & 4 \\
    \bottomrule
    \end{tabular}

    \vspace{-3mm}  

\end{table}

\begin{table}[h!]

    \centering
    \caption{\textbf{ST-SFCA model settings.}}  
    \label{tab:appendix_SFCA_settings2}
    
    \begin{tabular}{lp{40mm}}  
    \toprule
    Config & Value \\  
    \midrule
    input channels & 5 \\
    order & ``z", ``z-trans", ``hilbert", ``hilbert-trans" \\
    stride & [2, 2, 2, 2] \\
    encoder depths & [2, 2, 2, 6, 2] \\
    encoder channels & [32, 64, 128, 256, 512] \\
    encoder num head & [2, 4, 8, 16, 32] \\
    encoder patch size & [512, 512, 512, 512, 512] \\
    decoder depths & [2, 2, 2, 2] \\
    decoder channels & [64, 64, 128, 256] \\
    decoder num head & [4, 4, 8, 16] \\
    decoder patch size & [512, 512, 512, 512] \\
    mlp ratio & 4 \\
    \bottomrule
    \end{tabular}

    \vspace{-3mm}  

\end{table}

\subsection{Details of STA Module}
\label{appdx:Details_of_STA}
\subsubsection{Details of Attention Module}

The STA model employs multiple Attention modules with identical operations. The following section will describe their input-output characteristics and computational processes.

\vspace{8pt}
\noindent \textbf{Input and Output:}
Given the input tensors \( F_x \in \mathbb{R}^{B \times C \times N_x} \) and \( F_y \in \mathbb{R}^{B \times C \times N_y} \), where \( B \) is the batch size, \( C = 64\) is the number of channels, and \( N_x = N_y = 4096 \) is the sequence length of the inputs. The output tensor \( F'_x \) has the same size \( \mathbb{R}^{B \times C \times N_x} \), meaning the output dimensions match those of input \( F_x \).

\vspace{8pt}
\noindent \textbf{Computation Process:}
The queries \( Q \), keys \( K \), and values \( V \) are computed through convolutional layers:

\begin{equation}
Q, K, V = \left\{
\begin{aligned}
Q &= \text{Conv1d}_Q(F_y) \in \mathbb{R}^{B \times C \times N_y}, \\
K &= \text{Conv1d}_K(F_x) \in \mathbb{R}^{B \times C \times N_x}, \\
V_1 &= \text{Conv1d}_V1(F_x) \in \mathbb{R}^{B \times C \times N_x}, \\
V_2 &= \text{Conv1d}_V2(F_y) \in \mathbb{R}^{B \times C \times N_y}
\end{aligned}
\right.
\end{equation}
where Conv1d$(\cdot)$ is 1D convolutional layers. The attention matrix \( E \) is computed:

\begin{equation}
E = \frac{Q K^T}{\sqrt{C}} \in \mathbb{R}^{B \times N_y \times N_x}
\end{equation}
where \( C \) is the number of channels. \( E \) undergoes a fully connected transformation followed by ReLU activation:

\begin{equation}
E' = \text{ReLU}\left(\text{FC}_2\left(\text{ReLU}\left(\text{FC}_1(E)\right)\right)\right) \in \mathbb{R}^{B \times N_y \times N_x}
\end{equation}
where $\text{FC}_1$ and $\text{FC}_2$ are fully connected layers. The attention weights are computed using the Softmax function:

\begin{equation}
\text{Attention} = \text{Softmax}(E') \in \mathbb{R}^{B \times N_y \times N_x}
\end{equation}
The weighted sum is computed by applying the attention weights to the concatenated \( V \) tensors:

\begin{equation}
F'_x = \text{Attention} \times [V_2, V_1] \in \mathbb{R}^{B \times C \times N_x}
\end{equation}
where \( [V_2, V_1] \) denotes the concatenation of \( V_2 \)  and \( V_1 \) .

\subsubsection{Details of MLP Module}

The feature tensor \( F'_{st} \), obtained from the Spatial-Temporal Interaction, is subsequently passed through a Multi-Layer Perceptron (MLP) comprising fully connected layers. Initially, the feature tensor has dimensions of [4, 128, 4096], and after processing, the output feature tensor \( F_{sta} \) is generated with dimensions of [4, 4096, 128].

\subsection{Importance of Neighborhood Size in SFCA}  
\label{appdx:neighborhood_size}
 Tab.~\ref{tab: positional_encoding} further shows the importance of large neighborhood size in SFCA. Increasing the neighborhood size from 16 to 512 points significantly enhances the performance.

\begin{table}[h!]
\caption{\textbf{Patch size.} Effect of patch size changes on model performance, where P.S. represents the patch size in SFCA and Accuracy denotes the classification accuracy on the CIFAR10-DVS.}
\label{tab: positional_encoding}
\centering
\renewcommand{\arraystretch}{1.2}
\setlength{\tabcolsep}{3pt} 
\fontsize{9pt}{1pt}\selectfont
\begin{tabular}{@{}lccccccc@{}} 
\toprule
P.S. & 16 & 32 & 64 & 128 & 256 & \underline{\textbf{512}} & 1024 \\
\midrule
Accuracy (\%) & 82.9 & 83.0 & 83.8 & 84.0 & 84.8 & \underline{\textbf{85.2}} & 84.0 \\ 
\bottomrule
\end{tabular}
\end{table}

\subsection{Weight Configurations for \methodname\ and E2PNet}
\label{appdx:weight_configurations}
\begin{figure}[H]
\centering
\includegraphics[width = 0.5\textwidth]{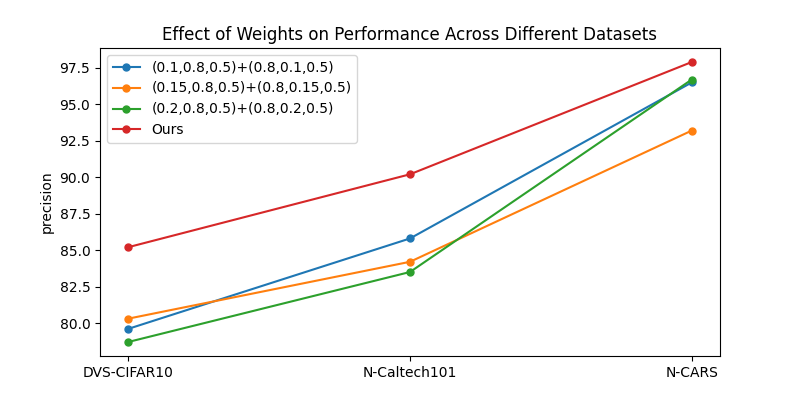}
   \caption{\textbf{\methodname\ vs E2PNet with 3 different scaling weight configurations.} The optimal weight configuration of E2PNet is different for different datasets. \methodname\ outperforms all of them without the need of weight tuning.}
\label{fig:weights}
\end{figure}

Fig.~\ref{fig:weights} shows that E2PNet requires different weight combinations to achieve optimal performance for different tasks/datasets. And even though with heavy weight tuning, E2PNet still performs worse than \methodname\ for all datasets.

\subsection{Visualization of Optical Flow Estimation}
\label{appdx:Visualization_flow}

\begin{figure}[!h]
  \centering
 {  
    \begin{minipage}{1\textwidth}  
      \centering
      \includegraphics[width=0.8\textwidth]{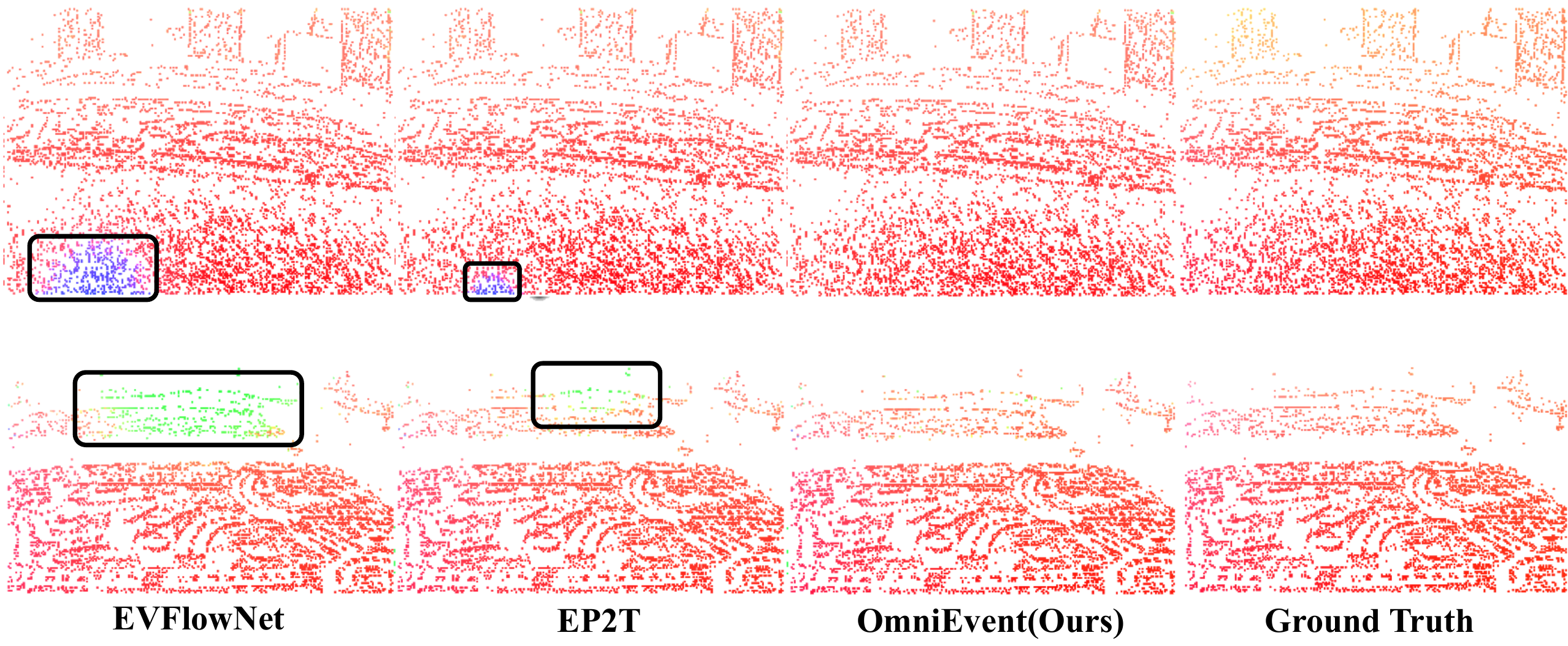}  
      \caption{\textbf{Qualitative comparisons of Optical Flow Estimation.} Optical flow visualization uses color encoding to represent motion information: hue indicates motion direction, with different colors representing different directions; saturation reflects motion magnitude, where higher saturation denotes stronger motion and lower saturation indicates weaker motion.}
      \label{fig:visualization_flow}
    \end{minipage}
  }
\end{figure}

As shown in Fig.~\ref{fig:visualization_flow}, our model significantly reduces the optical flow estimation error in event-dense regions compared to SOTA methods. This demonstrates that our proposed \methodname\ is more effective in preserving the spatiotemporal details in time-dense regions.

\clearpage
\end{document}